# A Critical Review of Large Language Models: Sensitivity, Bias, and the Path Toward Specialized AI

**Abstract:** This paper examines the comparative effectiveness of a specialized compiled language model and a general-purpose model like OpenAI's GPT-3.5 in detecting SDGs within text data. It presents a critical review of Large Language Models (LLMs), addressing challenges related to bias and sensitivity. The necessity of specialized training for precise, unbiased analysis is underlined. A case study using a company descriptions dataset offers insight into the differences between the GPT-3.5 and the specialized SDG detection model. While GPT-3.5 boasts broader coverage, it may identify SDGs with limited relevance to the companies' activities. In contrast, the specialized model zeroes in on highly pertinent SDGs. The importance of thoughtful model selection is emphasized, taking into account task requirements, cost, complexity, and transparency. Despite the versatility of LLMs, the use of specialized models is suggested for tasks demanding precision and accuracy. The study concludes by encouraging further research to find a balance between the capabilities of LLMs and the need for domain-specific expertise and interpretability.

**Keywords:** Large Language Models (LLMs), SDGs, Text Analytics, Comparative Study, Generative AI,

**Authors**
[Arash Hajikhani[1,2], Carolyn Cole[1]]

**Affiliations**
[1. Quantitative Science and Technology Studies, VTT Technical Research Centre of Finland, Tekniikantie 21, 02044 Espoo, Finland.
2. LUT University, School of Business and Management (LBM), Yliopistonkatu 34, 53850 Lappeenranta, Finland]

**Corresponding author(s)**
[Arash Hajikhani (arash.hajikhani@vtt.fi) - 'twitter: @Arash_Hajikhani']

## 1. Introduction

In the realm of Artificial Intelligence (AI), the rise of Large Language Models (LLMs) such as OpenAI's Generative Pretrained Transformer (GPT) series has introduced unprecedented capabilities in text summarization and classification (Min et al., 2021; Yoo et al., 2021). These AI juggernauts can dissect vast quantities of text, distill key points, and even classify documents with a level of speed and accuracy that leaves human ability far behind (Jiang et al., 2022). While we applaud these advancements, it's imperative to keep a clear perspective on their inner workings, particularly their training data and decision making procedures.

The advent of LLMs has undoubtedly revolutionized text analytics, but it has also introduced novel challenges concerning sensitivity and potential biases (Albrecht et al., 2022; Liang et al., 2021). Inherent in the training of these models is their susceptibility to embed the biases present in the training data, a subtle yet pervasive issue that can later be extremely difficult to detect and rectify (Alvi et al., 2019; Zhang & Verma, 2021). It's crucial, therefore, to scrutinize not only the LLMs themselves but also the mechanisms that train them. The broad and diverse nature of subjects that these models deal with, ranging from mundane queries to sensitive matters, necessitates a systematic and rigorous training approach.

Specialized language models that are trained meticulously, keeping the aforementioned factors in mind, can significantly reduce the risk of introducing biases and inaccuracies. Such models allow researchers to engage deeply with the training process, collecting appropriate data, performing diligent feature engineering, and fine-tuning the model's sensitivity to ensure that it is capable of handling the complexities of real-world texts.

To demonstrate the value of this specialized approach, we turn our focus to a case study involving the Sustainable Development Goals (SDGs) initiative. Established through the United Nations' 2030 Agenda in 2015, the SDGs provide a shared framework that guides stakeholders, from countries to corporations, in addressing pressing social, environmental, and economic challenges (Rosati & Faria, 2019; UN General Assembly, 2015; VNK, 2020).

The paper follows with a case study focused on the SDGs initiative, demonstrating the value of a specialized approach in SDG detection. The background section provides context for the study and highlights various interpretations and categorizations of SDGs. The methods section outlines the development process of the specialized SDG detection model and the experimental designs for comparing the performance of the GPT-3.5 model and the specialized model. The results section presents the findings from the comparative analyses, discussing the overlap and limitations of each model. The discussion section reflects on the implications of the observed differences and provides insights into the trade-offs between general and specialized models. Finally, the conclusion summarizes the main findings and reflects on the learnings from these experiments.

## 2. Background

Despite wide-spread adoption of 2030 Agenda, the SDGs lack unanimous interpretation, and some argue that the goals are too vague (Sianes et al., 2022; Spangenberg, 2017). Scholars have developed competing categorizations and indicators to enhance understanding of SDG application, suggesting a diversity of interpretation even among experts (Diaz-Sarachaga et al., 2018; Hametner & Kostetckaia, 2020; Lehtonen et al., 2016; Tremblay et al., 2020). Nonetheless, expert consensus has been successfully employed for the purpose of validating natural language processing (NLP) models built to automate SDG classification tasks (Guisiano et al., 2022). This demonstrates the potential of NLP models to achieve classification performance on par with experts, with the added benefit of reduced subjectivity bias.

In our endeavor to build a language model capable of identifying and understanding the nuances of the SDGs in text, we meticulously compiled a dataset and trained our model, keeping sensitivity and bias reduction as our primary targets. A detailed explanation of this process is given by Hajikhani et al. (2022). The culmination of this project was a rigorous experiment, deploying both our specialized SDG model and OpenAI's latest offering, GPT-3.5, on a selection of company text data. Our aim was to assess the models' sensitivity, comparing their detection and categorization of references to the SDGs in these texts. This formulates our primary comparative analysis. Given the divergent constructions and objectives of the specialized language model and the GPT, we would expect a fuzzy overlap in the classification results, where detections exhibit similarity for a core set of textual inputs but diverge when dealing with less evident detections in which the models' approach to nuance and context plays a larger role.

SDG detection presents a well-suited case for conducting this comparative study, as it reflects the complexities encountered in numerous real-world scenarios in which classifications are not mutually exclusive. Determining the relevant SDG(s) within a given piece of text can be intricate, requiring the consideration of nuance and contextual factors. Further, detections may differ between experts (or models) as these factors become more complex.

In addition to the primary analysis, we engage two supplementary analyses to further our understanding of GPT's artefact detection performance. The first supplementary analysis investigates the variation in GPT-3.5's own categorization performance when given a prescribed description in comparison with a description produced from its own capabilities. The second observes a short exercise wherein GPT-3.5's performance is evaluated using few-shot learning and a sample of observations taken from our specialized model's labelled training dataset. This exercise presents a view into the feasibility and suitability of few-shot learning as a mechanism for guiding the GPT-3.5 model towards more specified results.

In the following sections of this article, we delve into the details of this case study, offering insights into the challenges and successes encountered during the process. Through this analysis, we highlight the importance of an active role in training and developing specialized LLMs, showcasing how a thoughtful approach to AI development can lead to a more sensitive, unbiased, and accurate understanding of text.

## 3. Data

The data sample analyzed in the following analyses is used in conjunction with the ongoing INNOSDG project: Mapping Sustainable Development Activity; Its Evolution and Impact in Science, Technology, Innovation and Businesses. The project aims to operationalize big data approaches and create empirical tools to capture sustainable development activities resulting from Research and Development (R&D), public funding, or ecosystem collaboration.

The data sample is sourced from Finland's innovation policy agency's (Business Finland), public data bank (tietopankki). The sample is derived from a set of companies identified by Business Finland as young and ambitious. It totals 3,299 Finnish firms founded between 2009 and 2022. Prescribed company descriptions are sourced from three data providers: Vainu, CB Insights, and Pitchbook. Prescribed descriptions are available for 2,576 of the companies. Both the OpenAI GPT-3.5 model and the specialized SDG model were deployed on this set of company descriptions. From the GPT deployment, a classification result was returned for all prescribed descriptions. From the specialized SDG model deployment, 187 prescribed descriptions did not pass the model's text segment eligibility requirement. The final sample of companies eligible for the first comparison between the specialized SDG model and the GPT model totals 2,389.

For the second analysis focusing on GPT-3.5's performance on prescribed descriptions versus GPT description, firms founded after September 2021 were removed from the sample. OpenAI has noted that the GPT-3.5 model was trained on data through September 2021, so it has no basis for a GPT-based description of companies founded past this date. This results in a total sample of 2,550 for the second comparative analysis.

The final analysis testing the use of few-shot learning in GPT-3.5's categorization performance utilizes a small sample derived from the specialized model's labelled training dataset. This data is further described with the analysis, and a more detailed description can be found in Hajikhani et al. (2022).

## 4. Methods

In this section, we delineate our methodological approach for evaluating the performance of GPT against a specialized machine learning (ML) compiled model, focusing specifically on the detection of SDGs. We first elucidate the development process of the specialized SDG detection model, then elaborate on the strategies and experimental designs conceived to create benchmarking scenarios.

One avenue of experimentation involves contrasting the specialized SDG detection model against GPT's understanding of SDGs, which is invoked through multi-stage prompting. This is supplemented with an additional experiment that capitalizes on GPT's knowledge of companies. In practical terms, this implies querying the GPT for a company's description by providing the company's name as an input.

The next experimental design leverages our initial training data, originally used for constructing the specialized ML model, to exploit GPT's few-shot learning capability. The objective of this test was to feed labelled data into the GPT and evaluate its performance in assigning labels to unseen text.

Figure 1 provides a visual representation of our methodological pipeline, further illustrating our approach.

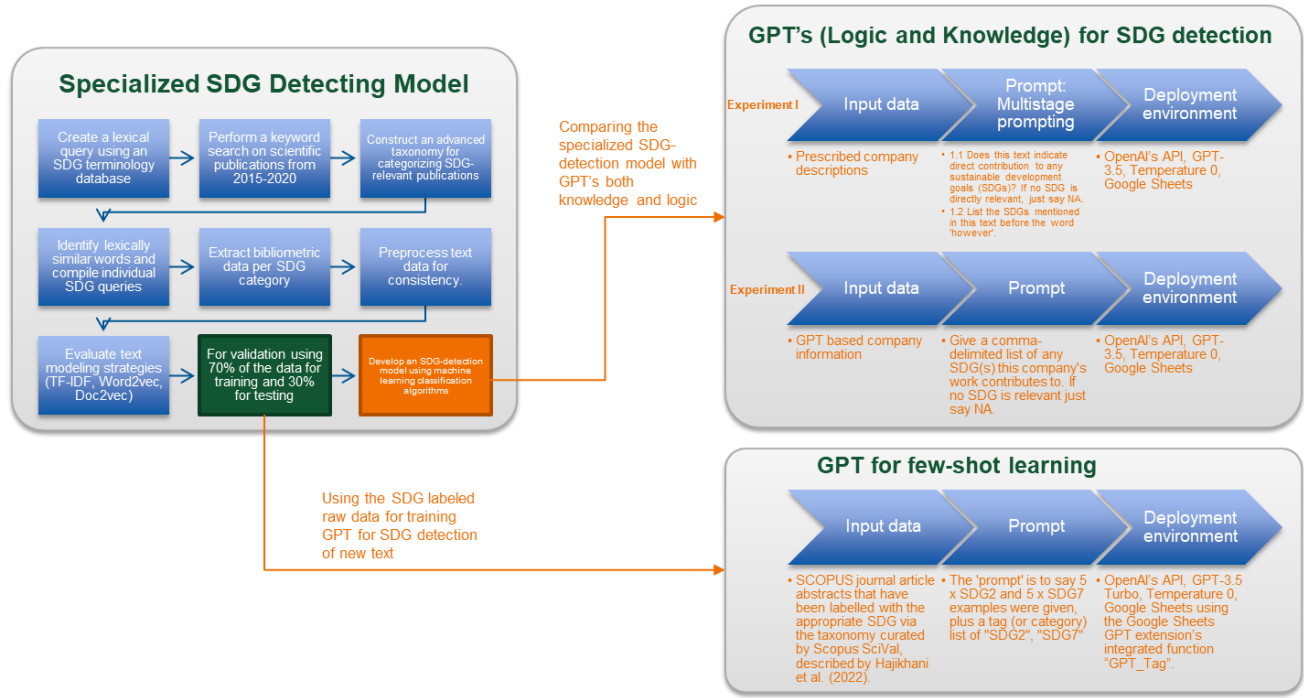

*Figure 1. Comparative Evaluation of GPT and Specialized SDG Detection Model*

## 4.1. Specialized SDG detecting model

The custom model was designed to discern SDGs annotations within science, technology, and innovation literature. The first stage entailed creating a lexical query utilizing an SDG terminology database. This led to a keyword search on scientific publications from 2015–2020, chosen due to its correlation with the 2030 SDG Agenda's initiation. This process produced publications pertinent to SDGs.

An advanced taxonomy, incorporating extant taxonomies and UN SDG document analyses, was constructed to categorize SDG-relevant publications. Lexically similar words were identified for each term, followed by compilation and search of individual SDG queries within the SCOPUS database. Bibliometric data were subsequently extracted per SDG category. The selected publications' titles and abstracts were utilized to train a model for automated detection of unseen SDG documents.

Machine learning (ML) classification algorithms were employed to develop an SDG-detection model. The identified SDG-related publications' text formed the classification algorithm's training dataset. Python was used for data structuring and ML model creation. Several classification methods were compared, using 70% of the data for training and the remaining 30% for testing.

Pre-classification text underwent preprocessing for consistency. Various text modeling strategies were evaluated, including Term Frequency Inverse Document Frequency (TF-IDF), Word2vec, and Doc2vec.

TF-IDF transformed text into a numerical representation, recognizing a term's significance within a document relative to other documents. Word2vec employed a neural network to generate a vector space model with contextually similar words in proximity. The skip-gram variant of Word2vec was used for improved accuracy with rare words. Doc2vec, an extension of Word2vec, provides representation for multiple documents. This was accomplished using Python's Gensim Word2vec feature, with the Google News corpus as a pre-trained model. The process can be seen in Figure 2.

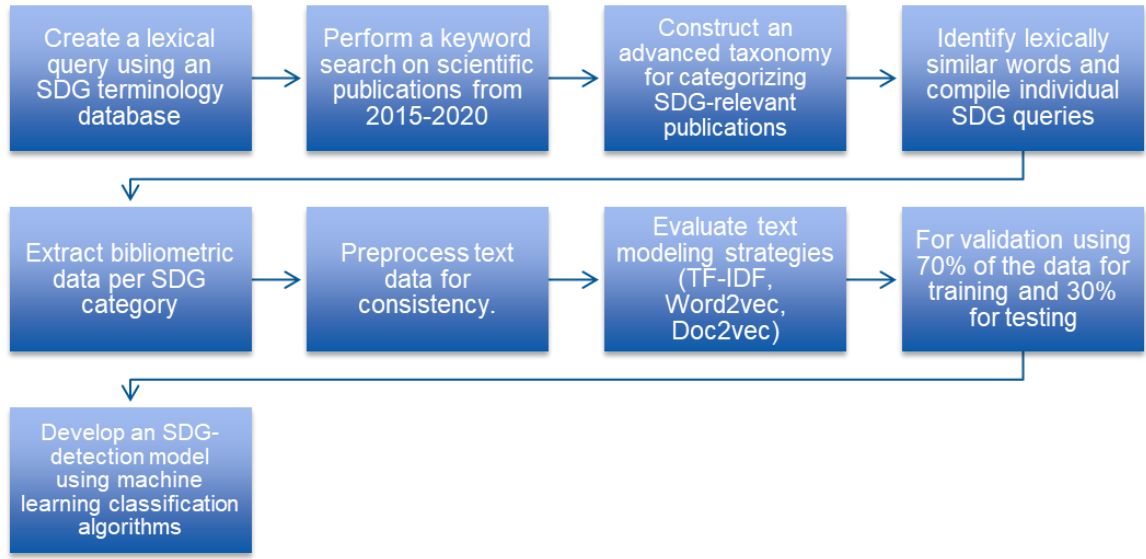

*Figure 2. Specialized SDG detection model development process*

## 4.2. GPT's (Logic and Knowledge) for SDG detection

Experiment I

The first experiment compares SDG detection by the GPT-3.5 model with the Specialized SDG model using prescribed company descriptions sourced from Vainu, CB Insights, and Pitchbook as previously described. This is followed by Experiment II, which focuses on the GPT model to explore its SDG detection when performed on a prescribed description versus GPT description.

The GPT-3.5 model was executed on the sample in April 2023. To facilitate this process, the model was deployed within the Google Sheets environment, utilizing OpenAI's integration with the platform. A connection was established with the OpenAI API using the required authentication key, and a given prompt was iterated over the provided set of descriptions. Each description, along with the prompt, was sent to the API for analysis, and the resulting responses were efficiently retrieved and stored in the same workbook as the original input. The limited length of the descriptions (median 73 words) meant that OpenAI's token limits were not a barrier for this particular task.

For analysis of the prescribed company descriptions, the model was implemented in two rounds using the prompts and specifications shown in Table 1.

*Table 1. Experiment I, prompt specification*

| Prompt step | Prompt | Value | Temp. | Token Limit | Model |
|---|---|---|---|---|---|
| 1.1 | Does this text indicate direct contribution to any SDGs? If no SDG is directly relevant, just say NA. | Prescribed company descriptions | 0 | No max | GPT-3.5 Turbo |
| 1.2 | List the SDGs mentioned in this text before the word 'however'. | Response to prompt 1.1 | 0 | No max | GPT-3.5 Turbo |

Both prompts were deployed using a temperature of 0 to minimize creativity, and no maximum token limit to allow for comprehensive response 'reasoning.' GPT-3.5 Turbo, the default OpenAI model, was utilized for all analyses. The initial prompt (1.1) requires direct SDG contribution in the text and provides an exit if no relevant SDG is detected to mitigate irrelevant classification. With no token limit, the model was given the freedom to elaborate on its SDG detection, enabling us to manually validate and sanity check a sample of responses. It was observed that the model occasionally indicated positive SDG contributions followed by negative ones, consistently separated with the word "however". To address this, a second prompt (1.2) was introduced as a simple cleaning mechanism, extracting positive SDGs mentioned only.

GPT-3.5 and the Specialized Model can both detect multiple SDGs from a single input. As a result, the models may simultaneously identify the same SDGs and different SDGs for a single company description. We use a non-restrictive intersection to identify overlaps in SDG detection, considering any common SDGs between the model results as an overlap. For example, from the same text input one model may detect SDGs 7 and 9 while the other model detects only SDG 7. Based on our chosen method, this is considered an overlap rather than a divergent result. This approach captures all overlaps, accounting for variations in sensitivity and lexical detection of each SDG by the models. The results can thus be considered an upper bound of common SDG detection by the two models.

### Experiment II

In this second experiment, our methodology involved harnessing the language model's capabilities to generate descriptions for a broad range of companies. This strategy provided us with an opportunity to evaluate the language model's competency in portraying corporate activities, a process informed by the model's comprehensive knowledge base derived from its training data. Subsequently, we deployed the same model to spotlight the SDG orientation within these company descriptions. This aspect of the methodology is based on the model's inherent comprehension and logical categorization of SDGs, which can potentially illuminate the presence and extent of SDG-aligned activities within each company's operational framework.

For implementation using the model's GPT description capabilities, only one prompt, specified in Table 2, was required. The response to this prompt relies on the model's capacity to recall and analyze information it has been exposed to during its training. It is assumed that the information regarding the companies in the sample would have been accessible on the open internet (via company websites, articles, or other online media) and incorporated into the model's training process. OpenAI has acknowledged that the model was trained on the open internet, though specific details regarding the training corpus have not been disclosed.

*Table 2. Experiment II, prompt specification*

| Prompt step | Prompt | Value | Temp. | Token Limit | Model |
|---|---|---|---|---|---|
| 1 | Give a comma-delimited list of any SDG(s) this company's work contributes to. If no SDG is relevant just say NA. | Company names | 0 | No max | GPT-3.5 Turbo |

Similar to Experiment I, a non-restrictive intersection is used to identify cases in which GPT-3.5 detects a common SDG between its deployment on prescribed company descriptions and its own GPT derived information.

### 4.3. GPT for few-shot learning

The methodology of this experiment consisted of assessing the performance of the GPT model under a few-shot learning scenario. The model utilized a labelled training dataset derived from SCOPUS journal article abstracts, each labelled according to the corresponding SDG using Scopus SciVal's taxonomy.

We derived a random sample of 200 observations from the original dataset of 31,998 entries. For this exercise, we selected two specific SDGs—SDG2 and SDG7— to test and utilized the "GPT_Tag" function of the Google Sheets GPT extension, providing ten examples (5 each of SDG2 and SDG7). The examples were selected randomly from the SDG2 and SDG7 stratifications of the original dataset, excluding the sample of 200 already selected. The GPT Tag function provided ideal deployment because it allows for user-defined classification tags and examples, guiding the model's outputs. Specification of the model parameters and prompts are in Table 3.

We set the temperature to 0, minimizing creativity, and used the GPT-3.5 turbo model for this analysis. We didn't enforce a token or a maximum number of tags limit. The model was then deployed against the sample of 200 abstracts.

*Table 3. GPT for few-shot learning experiment*

| Prompt step | Prompt | Value | Temp. | Token Limit | Model |
|---|---|---|---|---|---|
| 1 | 10 abstracts and their SDG labels were provided as examples (5 SDG2 and 5 SDG7). "SDG2, SDG7" were specified as the list of tags. | Journal abstract (label unseen) | 0 | No max | GPT-3.5 Turbo |

## 5. Results

This section first presents an analysis and comparison of the OpenAI GPT-3.5 model and the specialized model for SDG detection. A sample of company descriptions was used to examine the models' effectiveness in identifying SDGs. The results were also analyzed in terms of the quantity and relevance of the SDGs detected by each model. Further, a comparative assessment of SDG detection using the GPT-3.5 model was carried out on prescribed descriptions and generated descriptions. Lastly, we conducted an experiment using the GPT-3.5 model for few-shot learning on a labelled dataset to understand the model's capacity to detect based on limited examples. Each section within this chapter delves into the results and interpretations of these experiments, providing detailed insights into the strengths and limitations of the GPT-3.5 and specialized models for SDG detection.

## 5.1. Experiment I: Comparison of GPT-3.5 and Specialized Model for SDG Detection

The performance and capabilities of OpenAI GPT-3.5 and the Specialized SDG model in detecting SDGs from prescribed company descriptions were analyzed for a sample of 2,389 companies. Descriptive statistics are provided in Table 4. Of the total sample, 62.45% (1,492) showed an overlap between the two models, regardless of whether they detected any SDGs. Among the cases where both models detected at least one SDG, the overlap was only 10.46% (250 observations), indicating that the majority of overlaps occur where both models detected no SDGs.

*Table 4. OpenAI GPT-3.5 Prescribed Description vs Specialized Model*

| Statistic | Value | Percent of Total Companies |
|---|---|---|
| **Total Companies** | 2,389 | -- |
| Intersection: GPT-3.5 (Prescribed Description) vs Specialized Model | | |
| *including companies with no detected SDGs* | 1,492 | 62.45 |
| **Companies with Detected SDGs** | | |
| GPT-3.5 (Prescribed Description) | 1,019 | 42.65 |
| Specialized Model | 421 | 17.62 |
| Intersection: GPT-3.5 (Prescribed Description) vs Specialized Model | 250 | 10.46 |
| **Average Number of SDGs Detected per Company** | | |
| GPT-3.5 (Prescribed Description) | 1.74 | -- |
| Specialized Model | 1.12 | -- |

Figure 3 illustrates the SDG detection between the two models for companies where at least one SDG was detected by either model. OpenAI GPT-3.5 identified SDGs for 42.65% (1,019) of the companies (the blue circle on the left), while the specialized model detected SDGs for 17.62% (421) of the companies (the red circle on the right). Although the GPT model had a higher detection rate, this does not necessarily imply better performance in identifying SDGs. The brevity and generality of company descriptions (median 73 words) limit the opportunity for SDG-related text to be present. Unless an SDG is highly relevant to a company's activity, it is unlikely to be detected in such a concise statement. SDG detection by the specialized model indicates a high level of reliability, although with a conservative detection rate. In contrast, the GPT model casts a wider net, detecting SDGs in more than twice the number of company descriptions. This suggests a more liberal interpretation of SDG classification by the GPT model, which may also dilute the meaningfulness of detections.

Returning to Table 4, both models can detect multiple relevant SDGs for a given input text. On average, OpenAI GPT-3.5 identified 1.74 SDGs per description, while the specialized model detected 1.12 SDGs. This aligns with the GPT model's liberal identification approach – along with detecting SDGs in a broader range of descriptions, the GPT model also identifies more SDGs per description.

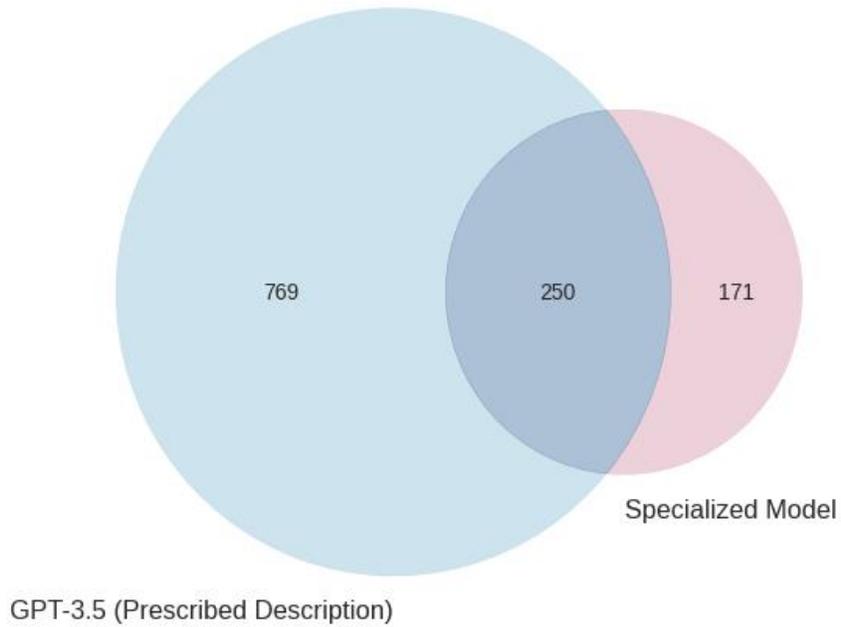

***Figure 3. Companies Detected as SDG Relevant: OpenAI GPT-3.5 (Prescribed Description) vs. Specialized Model***

Figure 4 shows each SDG's detection rate, representing the percentage of descriptions in which the SDG is detected, between the two models. The rates differ significantly in magnitude, reflecting the higher detection rates by the GPT model. However, the distributions are relatively similar and share two out of three top detected SDGs. The top three SDGs detected by the GPT model in this sample are SDGs 9, 4, and 3, while the top three from the specialized SDG model are SDGs 9, 3, and 12.

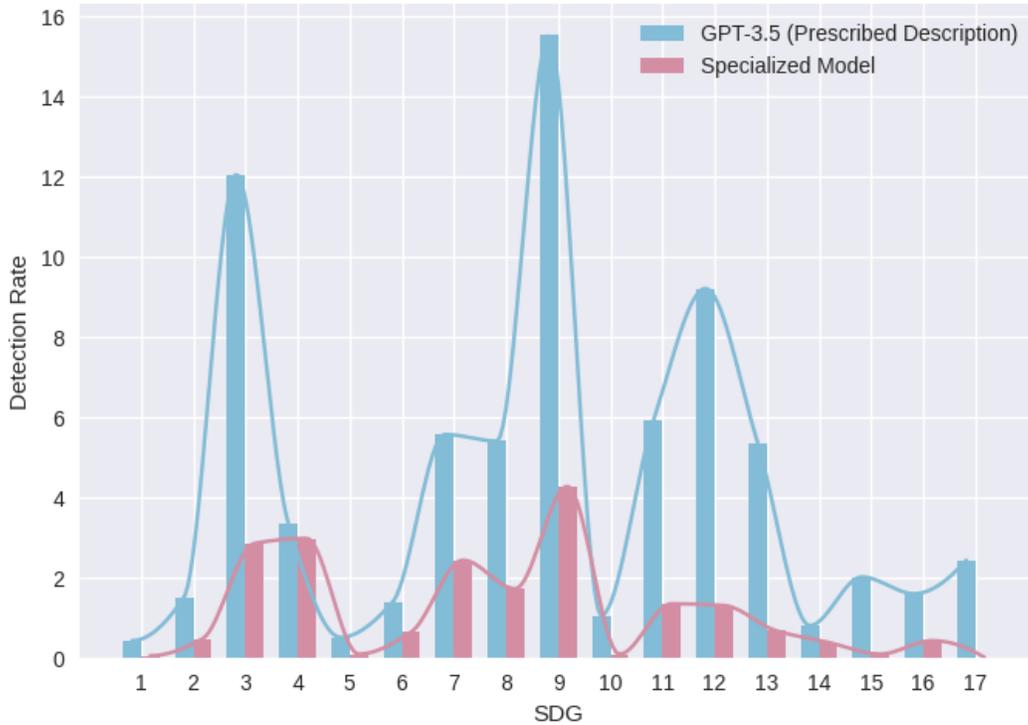

**Figure 4. Detection Rate by SDG and Model: OpenAI GPT-3.5 (Prescribed Description) vs. Specialized Model**

Overall, the results suggest that the specialized model exhibits a more conservative and robust approach to detecting SDGs among the analyzed companies. With a lower percentage of companies (17.62%) and a lower average number of SDGs (1.12) detected per company, the specialized model focuses on more relevant and specific indicators of SDGs. On the other hand, OpenAI GPT-3.5 demonstrates a more liberal approach, identifying SDGs for a higher percentage of companies (42.65%) and detecting a higher average number of SDGs (1.74) per company. While this may indicate a broader coverage, it also suggests the possibility of including less relevant or tangential information related to SDGs.

When choosing between these models, it's important to consider the specific context and purpose of SDG detection. The specialized model prioritizes precision, focusing on highly relevant SDGs, resulting in a more reliable and focused assessment. On the other hand, OpenAI GPT-3.5 provides a broader analysis which could be useful for exploratory purposes as it captures a wider range of information. However, a caveat is that this may include both highly relevant and minimally relevant SDGs.

## 5.2. Experiment II: Comparison of GPT-3.5 SDG Detection on Prescribed Description vs GPT Description

Table 5 presents the comparison of OpenAI GPT-3.5's SDG detection on prescribed description and GPT generated description for a sample of 2,550 companies. Of the total sample, 81.10% (2,086) show an overlap between the prescribed description and GPT description, regardless of whether an SDG was detected. 40.71% (1,038) of companies had SDGs identified through the prescribed description approach, while 48.27% (1,231) had SDGs detected through the GPT description. Figure 6 shows that the distribution of SDGs detected between the two approaches is also fairly similar, though overall magnitudes differ.

*Table 5. OpenAI GPT-3.5: Prescribed Description vs GPT Description*

| Statistic | Value | Percent of Total Companies |
|---|---|---|
| **Total Companies** | 2,550 | -- |
| Intersection: Prescribed Description vs GPT Description *including companies with no detected SDGs* | 2,086 | 81.10 |
| **Companies with Detected SDGs** | | |
| Prescribed Description | 1,038 | 40.71 |
| GPT Description | 1,231 | 48.27 |
| Intersection: Prescribed Description vs GPT Description | 890 | 34.90 |
| **Average Number of SDGs Detected per Company** | | |
| Prescribed Description | 1.73 | -- |
| GPT Description | 2.89 | -- |

A large number of overlaps occur where both approaches detected SDGs, with 890 companies having at least one SDG in common between the two approaches. This overlap, focusing on the companies for which SDGs were detected, is depicted in Figure 5.

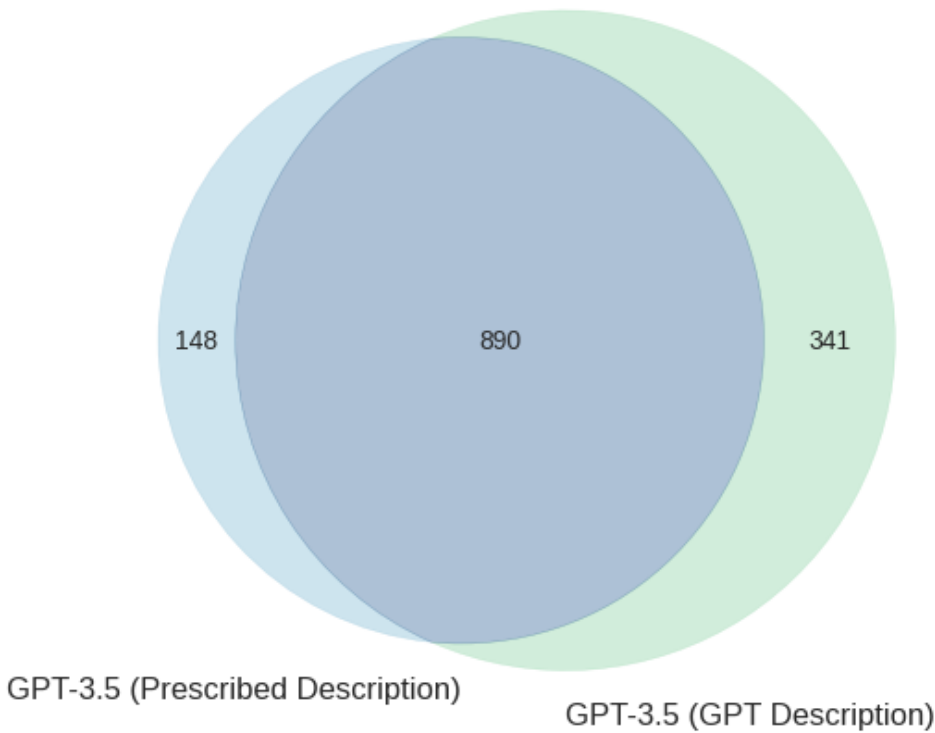

*Figure 5. Companies Detected as SDG Relevant: OpenAI GPT-3.5 (Prescribed Description) vs. OpenAI GPT-3.5 (GPT Description)*

The average number of SDGs detected per company is 1.73 for the prescribed description and 2.89 for the GPT description. There are two possible explanations for this substantial difference. On one hand, the approach that uses GPT description, which likely relies on a broader range of company information beyond the prescribed description, may result in a higher number of SDGs detected per company because it has access to a greater quantity of information about the company. This would indicate a potentially more comprehensive analysis. Alternatively, given a broader base of information to draw from for the company, the GPT description approach may be detecting a greater number of SDGs with low or tangential relevance to the companies' activity, due to the liberal tendency of the GPT-3.5 model's SDG detection.

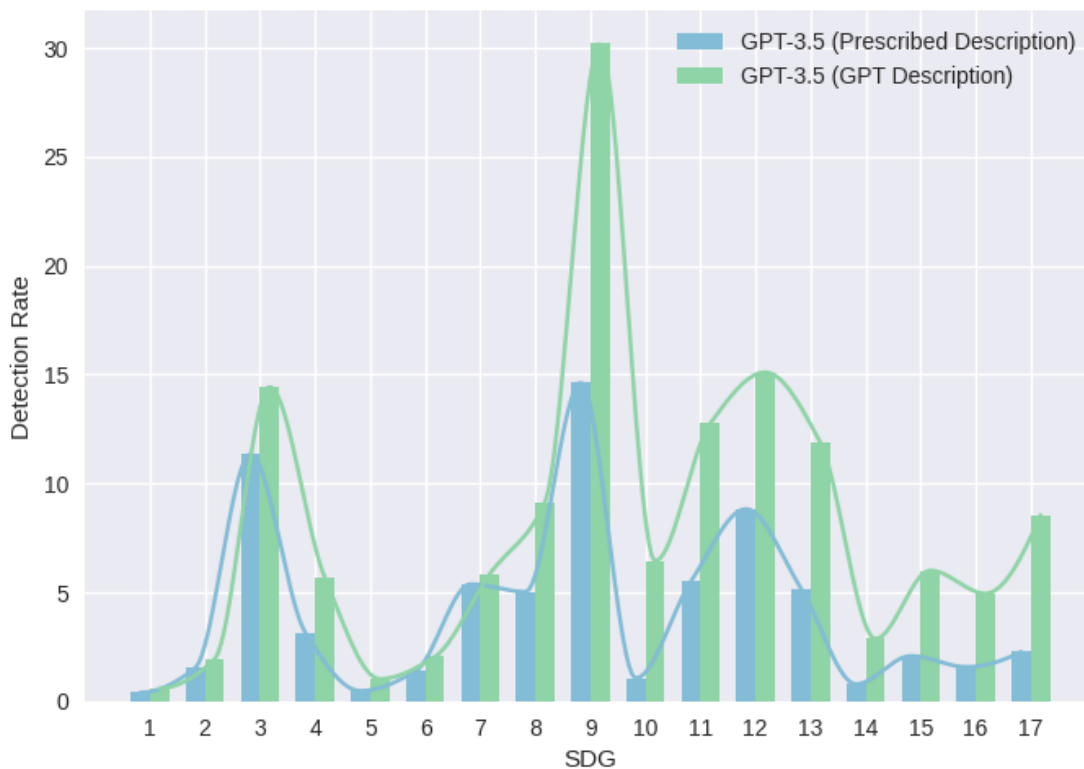

**Figure 6. Detection Rate by SDG and Model: OpenAI GPT-3.5 (Prescribed Description) vs. OpenAI GPT-3.5 (GPT Description)**

Figure 6 shows that the distribution of SDGs detected between the two approaches is fairly similar, though overall magnitudes differ.

Overall, the prescribed description and GPT description approaches have a considerable intersection and perform similarly in terms of coverage. However, there is substantial difference in the average number of SDGs detected per company when using the GPT description, which exhibits a much higher average number of SDGs per company compared to detection from the prescribed description.

### 5.3. GPT for few-shot learning exercise

In addition to the GPT deployments for this overlap analysis, a separate exercise was performed to investigate GPT's performance under few-shot learning using data from the specialized model's labelled training dataset. This dataset is comprised of SCOPUS journal article abstracts that have been labelled with the appropriate SDG to which the article relates via the taxonomy curated by Scopus SciVal. The data is described in greater detail by Hajikhani et al. (2022). For this exercise, a random sample of 200 observations was selected from the labelled training dataset of 31,998 observations. The distribution of SDG labels across this sample of 200 is shown in Table 6 (N).

For this analysis, the Google Sheets GPT extension's integrated function "GPT_Tag" was utilized due to its suitable construction. The Tag function includes in its list of parameters a user-defined set of classification tags and a table of examples. This allows the user to pass a list of specified outputs, as well as a set of expected input-output pairs, to the model for consideration in its evaluation of the input value. Due to the limited number of tokens GPT allows per API call and the average length of abstracts in the data sample (avg. 218 words), a maximum of ten examples were permissible per call. This precluded the ability to provide the model with one example per SDG, much less multiple examples per SDG. As a result, only two SDGs were chosen for this test – SDG2 and SDG7. The list of tags specified was limited to "SDG2, SDG7" with the expectation that this would confine the model's output to only tag these SDGs for which it was provided examples. Ten observations (5 x SDG2 and 5 x SDG7) were randomly selected from the full training data (excluding the 200 observations already selected for the validation sample) to populate the table of examples. In the model's deployment, a temperature of 0 was utilized to minimize creativity, no token limit was enforced, no maximum number of tags were enforced in the output (to allow for both SDG2 and SDG7 to be tagged simultaneously), and the GPT-3.5 turbo model was used. The analysis was run in June 2023. Table 6 displays the analysis results.

*Table 6. GPT Model Performance in SDG Label Identification*

| Label (1) | N (2) | E(y) (3) | Total identification y Count (4) | % (5) | Identification as expected y = E(y) Count* (6) | % (7) | Correct identification regardless of E(y) y=Label Count (8) | % (9) |
|---|---|---|---|---|---|---|---|---|
| SDG1 | 12 | 0 | 6.00 | 50.00 | [4] | 33.33 | 4 | 33.33 |
| SDG2 | 15 | 15 | 60.00 | 400.00 | 15 | 100.00 | 15 | 100.00 |
| SDG3 | 11 | 0 | 14.00 | 127.27 | [6] | 54.55 | 5 | 45.45 |
| SDG4 | 16 | 0 | 5.00 | 31.25 | [11] | 68.75 | 3 | 18.75 |
| SDG5 | 11 | 0 | 6.00 | 54.55 | [6] | 54.55 | 4 | 36.36 |
| SDG6 | 15 | 0 | 14.00 | 93.33 | [3] | 20.00 | 9 | 60.00 |
| SDG7 | 10 | 10 | 32.00 | 320.00 | 10 | 100.00 | 10 | 100.00 |
| SDG8 | 12 | 0 | 6.00 | 50.00 | [3] | 25.00 | 3 | 25.00 |
| SDG9 | 9 | 0 | 1.00 | 11.11 | [5] | 55.56 | 0 | 0.00 |
| SDG10 | 18 | 0 | 5.00 | 27.78 | [12] | 66.67 | 0 | 0.00 |
| SDG11 | 7 | 0 | 5.00 | 71.43 | [0] | 0.00 | 0 | 0.00 |
| SDG12 | 9 | 0 | 3.00 | 33.33 | [1] | 11.1 | 2 | 22.22 |
| SDG13 | 13 | 0 | 16.00 | 123.08 | [2] | 15.38 | 5 | 38.46 |
| SDG14 | 14 | 0 | 15.00 | 107.14 | [2] | 14.29 | 7 | 50.00 |
| SDG15 | 16 | 0 | 5.00 | 31.25 | [5] | 31.25 | 0 | 0.00 |
| SDG16 | 12 | 0 | 2.00 | 16.67 | [5] | 41.67 | 1 | 8.33 |
| Total | 200 | 200 | 195.00 | 67.50 | 90 | 45.00 | 68 | 34.00 |

Note: Multiple SDGs can be identified per abstract. An average of 1.44 SDGs is identified per abstract.
*Counts shown in brackets represent abstracts in which the model identified 0 SDGs, as expected.

As described, a random sample of 200 observations selected from the Specialized model's training set forms the validation data of this analysis. The distribution of SDG labels across this sample is given in column 2 and the expected model output in column 3. Because we limited the tags and examples to SDGs 2 and 7, we expect the model to give 0 tags for the abstracts associated with the remaining SDGs. However, the model does not behave in this manner. Despite the tag list not including the other 15 SDGs, the model identified these SDGs (except for SDG17) in its output anyway. This suggests that the model disregards the limitation and employs its knowledge of the SDGs not just on the tags specified, but further extrapolates to tag unlisted SDGs as well.

Total identification (correct or incorrect) shows extreme over identification in the two SDGs for which examples were provided. However, these are not the only SDGs that experience overidentification, as SDGs 3, 13, and 14 are also overidentified. Meanwhile, SDGs 9 and 16 show the most under identification. While 195 SDGs are identified overall, the model only identifies SDGs for 67.5% of the abstracts. In some cases, the model identified multiple SDGs per abstract, with an overall average of 1.44 SDGs identified per abstract, of those with an identified SDG.

When accounting for abstracts in which we would expect the model to return no SDGs (i.e., abstracts labelled with SDGs other than 2 and 7), we observe that 45% of the model's output is in line with expectation. Importantly, the model detects SDGs 2 and 7 in 100% of the abstracts with those SDG labels. This suggests a very high capture rate with few-shot learning. Note however that overidentification was also extremely high in these two SDGs.

Finally, despite the limitation to SDG2 and SDG7 in the tag list and table of examples provided to the model, it does identify the remaining SDGs, as well. However, these are largely misidentifications. Notably, by comparing the total identifications (columns 4 and 5) with the correct identifications (columns 8 and 9) of SDGs 9, 10, 11, and 16, we can see that the model did not identify the correct SDG in any of these cases. Other SDGs have a better rate of correct detection, for example SDG6 (60%) and SDG14 (50%). Overall, 34% of the abstracts are identified with the correct label.

## 6. Discussion

The comparison between a specialized SDG classification model and OpenAI GPT-3.5 elucidates notable differences in SDG detection performance. The specialized SDG model is specifically trained and tuned to detect relevant SDGs from the input text with precision and reliability, thus revealing a more robust but conservative approach. On the other hand, OpenAI GPT-3.5, as a general language model, is more liberal, identifying SDGs in a broader spectrum of company descriptions.

A significant point of deviation between these models is the apparent difference in specificity of SDG detection, as observed in the analysis. The OpenAI GPT-3.5 model, with its general training, leans towards detecting a higher number of SDGs per description and in a higher percentage of companies overall. This reflects the model's ability to capture a range of information, but at the same time might lead to less meaningful detection due to the more liberal application of SDG relevance.

Meanwhile, the specialized SDG model, by nature of its focused training, detected fewer SDGs in company descriptions. This is an indication of the model's conservative approach, limiting detections to SDGs that are highly relevant to the company's activities. The specificity in the model's detection suggests that the detected SDGs are likely to be more meaningful and significant in relation to the company's activity.

When working with the OpenAI GPT-3.5 model, it's important to consider not only the model's analysis capabilities, but also the input on which it is exercised. GPT-3.5 has been trained on a vast amount of data, granting it an impressive ability to retrieve, summarize, and analyze information. However, specific details about the training data have not been publicly disclosed. This lack of transparency makes it challenging to determine the sources from which the GPT description is based. This becomes even more of a black box when the information requested from the GPT model does not have abundant or clear sources for spot check verification (such as details of a young startup). The GPT description approach using the GPT-3.5 model yielded substantially different results in terms of detection rate than the prescribed description approach. This may be seen as wider detection ability, but it should also highlight the need for caution when relying on GPT generated content for analyses requiring focused and robust detection.

Our final experiment presents thought-provoking considerations for the use of GPT with few-shot learning. A key takeaway of this experiment lies in the limitations it reveals for few-shot learning with GPT. There are considerable limits to the use of few-shot learning with GPT due to the model's constraints. The API used to access the GPT model has a restricting token limit. Considering the average length of abstracts in our dataset, this confined us to a maximum of ten examples per interaction. This imposed a significant constraint on our ability to test the model with few-shot learning, as it prohibited the presentation of one example per SDG, let alone multiple instances per goal. As a result, it was not possible to evaluate the GPT model's capability to generalize from a handful of examples to the full array of SDGs in a controlled manner. Despite our adjustment limiting the experiment to two SDGs, the model heavily extrapolated and categorized 16 SDGs in the test, with varying degrees of accuracy. This suggests that restrictions placed on the model through available channels may not be reliably binding.

The use of few-shot learning with GPT as an avenue for task adaptation or more specialized performance should be carefully considered. Our experiment demonstrates the limitations of this method for use cases involving long input or a high number of classification categories. Further, unexpected results may be difficult to interpret given the relative obscurity of the model's procedures. While this method may be useful under certain conditions, it does not replace the abilities of a specialized language model with task-specific training.

Our methodological contribution involves a comparative analysis between GPT-3.5 and the specialized SDG model, providing valuable insights into the strengths and weaknesses of both general and task-specific AI models. We not only assessed performance metrics but also scrutinized their usability, interpretability, and limitations.

Furthermore, the method of juxtaposing a general model's performance with a specialized model, considering differences in training data and potential biases, adds another dimension to our understanding of AI capabilities. This study allows us to contemplate the potential of large language models beyond their initial training objectives, providing insights into how we can more effectively leverage their capabilities.

Lastly, our analysis evaluating both false positives and false negatives presents a holistic picture of the models' capabilities and highlights potential areas for future research. This could prove instrumental in the fine-tuning of these models or even in the development of hybrid models that blend the strengths of both general and specialized models.

## 7. Conclusion

The observed deviation between the Specialized SDG model and OpenAI's GPT-3.5 underscores the need for careful consideration in the application of these models. It accentuates the trade-off between the vast coverage of general models like GPT-3.5 and the precision of specialized models. This disparity, resulting from differences in training data and model parameter tuning capabilities, warrants thoughtful contemplation. Importantly, despite some observed overlap in the results produced by large language models (LLMs) like GPT-3.5 and specialized models, their performance should not be considered interchangeable. The choice between a general or specialized model must be dictated by the specific requirements of a given task. Broad, catch-all classification tasks may benefit from LLMs, whereas precision-focused tasks necessitate the use of specialized models.

While the progression of LLMs suggests the potential for more nuanced SDG detection in the future, it is contingent upon researchers engaging in more specific data training and further model parameter tuning. This approach could allow LLMs to match the precision of specialized models without compromising their broad coverage. However, this observation comes with a vital cautionary note. It is clear that LLMs, such as GPT-3.5, operate as black-box models, leaving us without a clear view of how they arrive at their conclusions. Consequently, their expansive, liberal application could lead to unpredicted and, in some cases, undesired outcomes. As such, a more reliable approach, especially when accuracy and transparency are of utmost importance, would be to utilize a specialized model tailored to the task at hand.

Our experiment extends beyond the specific comparison of GPT-3.5 and the specialized model. It offers a unique vantage point on contrasting a highly specialized machine learning model with an autonomous LLM. It invites scholars to carefully consider the trade-offs of using LLMs, including their cost, complexity, and opacity. It also underlines that for many applications, developing a specialized model tailored to the task at hand might be more straightforward, cost-efficient, and transparent. While LLMs are undoubtedly powerful and versatile, their use should not be considered a one-size-fits-all solution. Researchers and practitioners are encouraged to explore other alternatives, such as compiling and training a machine learning model on their own data. In this light, our study underlines that there is no universal answer, and the choice of the model should be dictated by the task, the data, and the specific requirements of each use case.